\documentclass[sigconf]{aamas}

\usepackage{balance} %
\usepackage{xspace}
\usepackage{cleveref}
\usepackage{algorithm}
\usepackage{algpseudocode}
\usepackage{tcolorbox}
\usepackage[dvipsnames]{xcolor}
\usepackage{enumitem}
\usepackage{multirow}
\usepackage{makecell}
\usepackage{subcaption}

\newcommand*\circled[1]{\tikz[baseline=(char.base)]{
            \node[shape=circle,draw,fill=blue, text=white,inner sep=2pt] (char) {#1};}}

\setcopyright{ifaamas}
\acmConference[AAMAS '26]{Proc.\@ of the 25th International Conference
on Autonomous Agents and Multiagent Systems (AAMAS 2026)}{May 25 -- 29, 2026}
{Paphos, Cyprus}{C.~Amato, L.~Dennis, V.~Mascardi, J.~Thangarajah (eds.)}
\copyrightyear{2026}
\acmYear{2026}
\acmDOI{}
\acmPrice{}
\acmISBN{}

\acmSubmissionID{1491}

\title{UNCAP: Uncertainty-Guided Neurosymbolic Planning \\Using Natural Language Communication for \\Cooperative Autonomous Vehicles}

\author{Neel P. Bhatt*}
\affiliation{
  \institution{The University of Texas at Austin}
  \city{Austin, Texas}
  \country{United States}}
\email{npbhatt@utexas.edu}

\author{Po-han Li*}
\affiliation{
  \institution{The University of Texas at Austin}
  \city{Austin, Texas}
  \country{United States}}
\email{pohanli@utexas.edu}

\author{Kushagra Gupta*}
\affiliation{
  \institution{The University of Texas at Austin}
  \city{Austin, Texas}
  \country{United States}}
\email{kushagrag@utexas.edu}

\author{Rohan Siva}
\affiliation{
  \institution{The University of Texas at Austin}
  \city{Austin, Texas}
  \country{United States}}
\email{rohansiva@utexas.edu}

\author{Daniel Milan}
\affiliation{
  \institution{The University of Texas at Austin}
  \city{Austin, Texas}
  \country{United States}}
\email{dm10787@utexas.edu}

\author{Alexander T. Hogue}
\affiliation{
  \institution{The University of Texas at Austin}
  \city{Austin, Texas}
  \country{United States}}
\email{alex.hogue@utexas.edu}

\author{Sandeep P. Chinchali}
\affiliation{
  \institution{The University of Texas at Austin}
  \city{Austin, Texas}
  \country{United States}}
\email{sandeepc@utexas.edu}

\author{David Fridovich-Keil}
\affiliation{
  \institution{The University of Texas at Austin}
  \city{Austin, Texas}
  \country{United States}}
\email{dfk@utexas.edu}

\author{Zhangyang Wang}
\affiliation{
  \institution{The University of Texas at Austin}
  \city{Austin, Texas}
  \country{United States}}
\email{atlaswang@utexas.edu}

\author{Ufuk Topcu}
\affiliation{
  \institution{The University of Texas at Austin}
  \city{Austin, Texas}
  \country{United States}}
\email{utopcu@utexas.edu}

\begin{abstract}
Safe large-scale coordination of multiple cooperative connected autonomous vehicles (CAVs) hinges on communication that is both efficient and interpretable.
Existing approaches either rely on transmitting high-bandwidth raw sensor data streams or neglect perception and planning uncertainties inherent in shared data, resulting in systems that are neither scalable nor safe.
To address these limitations, we propose \textbf{U}ncertainty-Guided \textbf{N}atural Language \textbf{C}ooperative \textbf{A}utonomous \textbf{P}lanning (\oursnospace), a vision–language model-based planning approach that enables CAVs to communicate via lightweight natural language messages while explicitly accounting for perception uncertainty in decision-making. \ours features a two-stage communication protocol: (i) an ego CAV first identifies the subset of vehicles most relevant for information exchange, and (ii) the selected CAVs then transmit messages that quantitatively express their perception uncertainty. By selectively fusing messages that maximize mutual information, this strategy allows the ego vehicle to integrate only the most relevant signals into its decision-making, improving both the scalability and reliability of cooperative planning. Experiments across diverse driving scenarios show a $63\%$ reduction in communication bandwidth with a $31\%$ increase in driving safety score, a $61\%$ reduction in decision uncertainty, and a four-fold increase in collision distance margin during near-miss events.

\end{abstract}

\keywords{Natural Language Communication, Planning Under Uncertainty, Cooperative Autonomous Vehicles, Vision-Language Models, Safety and Efficiency.
}

\newcommand{\BibTeX}{\rm B\kern-.05em{\sc i\kern-.025em b}\kern-.08em\TeX}

\newcommand{\ours}{\texttt{UNCAP}\xspace}
\newcommand{\bare}{\texttt{BARE}\xspace}
\newcommand{\spare}{\texttt{SPARE}\xspace}

\newcommand{\oursnospace}{\texttt{UNCAP}}
\newcommand{\barenospace}{\texttt{BARE}}
\newcommand{\sparenospace}{\texttt{SPARE}}
\newcommand{\langcoop}{\texttt{LangCoop}\xspace}
\newcommand{\nocomm}{\texttt{No-Comm}\xspace}
\newcommand{\cav}{\textrm{cav}}
\newcommand{\vlm}{\textrm{vlm}}
\newcommand{\ego}{\textrm{ego}}
\newcommand{\goal}{\textrm{goal}}

\definecolor{datasetAcolor}{HTML}{A3C4F3} %
\definecolor{datasetBcolor}{HTML}{FBC4AB} %
\definecolor{datasetCcolor}{HTML}{C9E4DE} %

\begin{document}

\pagestyle{fancy}
\fancyhead{}

\maketitle 
\renewcommand{\thefootnote}{\fnsymbol{footnote}}
\footnotetext[1]{These authors contributed equally to this work.}
\renewcommand{\thefootnote}{\arabic{footnote}}

\section{Introduction}

\begin{figure*}[t] %
    \centering
    \includegraphics[width=0.7\textwidth, trim={0cm, 0cm, 1.6cm, 0cm}, clip]{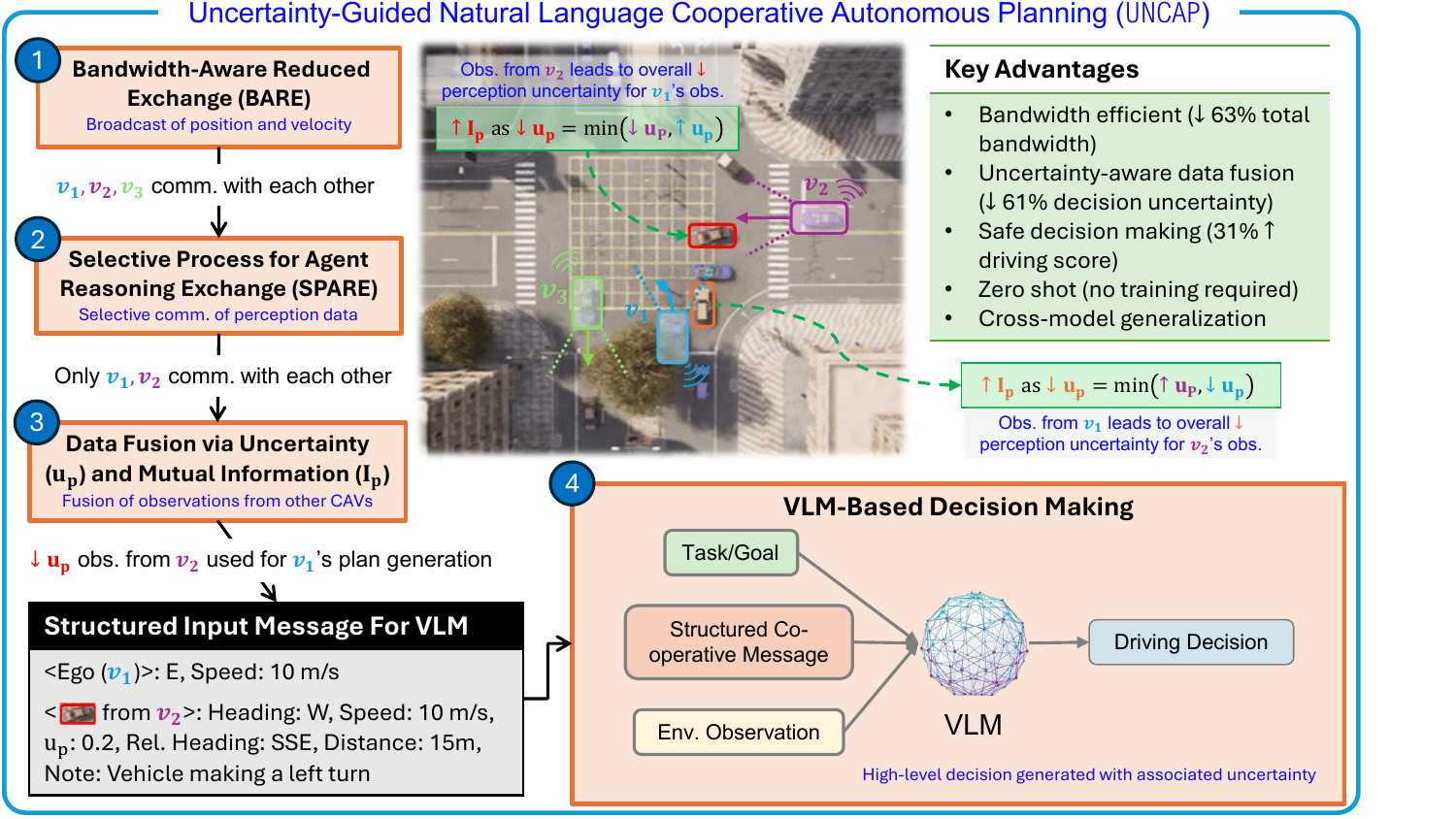}
    \caption{Overview of \oursnospace. To minimize bandwidth, vehicles ($v_1, v_2, v_3$) first engage in \barenospace~and share essential state information. \spare then enables selective communication among relevant agents ($v_1, v_2$) to focus on critical interactions. Observations are fused based on their contribution to reducing perception uncertainty ($u_p$), prioritizing information that maximizes mutual information ($I_p$) for the ego vehicle. The fused messages are then put in a structured format and used by a VLM to produce driving decisions with an associated decision uncertainty score ($u_d$) for safe and interpretable planning.}
    \Description[<short description>]{}
    \label{fig:framework}
\end{figure*}

We study cooperation among connected autonomous vehicles \allowbreak (CAVs) using \emph{natural language} communication as the medium of mutual interaction. While inter-vehicle cooperation has been extensively studied, most existing approaches focus on transmitting raw sensor data or features extracted from deep neural networks \cite{xu2022v2x,gao2018object,opv2v}. These modalities require substantial communication bandwidth, impose high computational costs for inference, and implicitly assume homogeneous sensing and processing capabilities across cooperative vehicles. Although a few recent works have explored cooperative communication through natural language in CAVs equipped with vision–language models (VLMs), these methods are typically restricted to pairwise interactions and do not explicitly account for uncertainties in vehicle perception, raising the question of how language-based communication and planning frameworks can scale to larger CAV fleets while accounting for perception uncertainties \cite{gao2025langcoop,sha2025languagempclargelanguagemodels, cui2025towards}.

The use of VLMs opens new opportunities for cooperative autonomous driving. By bridging raw perceptual inputs with high-level reasoning, VLMs allow vehicles to interpret their surroundings, explain their behaviors, and incorporate human feedback. In particular, they enable CAVs to exchange semantic information through text rather than raw sensor data, supporting scalable, low-bandwidth coordination. For instance, a vehicle might report \emph{``pedestrian entering crosswalk at 3 m, speed 1.2 m/s''} or broadcast \emph{``changing lane left in 2 s''}, allowing nearby planners to adjust trajectories and avoid conflicts. Moreover, VLMs can enrich such messages with confidence scores and relevance tags, so that downstream planning modules selectively integrate only the messages that reduce decision uncertainty.

Beyond coordination, VLMs can adapt to human driving preferences and traffic regulations expressed in natural language, e.g. \emph{``maintain a safe distance on wet roads"} or \emph{``drive more conservatively near schools"}. This capability enhances both flexibility and interpretability in autonomous driving. Semantic natural language messages also reduce transmission latency relative to full sensor streams, preserve privacy by avoiding the exchange of raw images, and provide interpretable explanations that can be audited by human operators and regulators.

Natural language has been previously explored as a medium for collaborative autonomous driving in works such as \langcoop \cite{gao2025langcoop,cui2025towards}. However, these frameworks face key challenges in scalability and robustness for real-world deployment.
First, they evaluate performance primarily with traditional metrics such as driving score or time-to-collision, which reflect driving quality but fail to capture perception and planning certainty and bandwidth constraints, all of which are crucial for decision-making in autonomous driving.
Second, communication is restricted to pairwise exchanges; this scales poorly to larger fleets and risks redundancy or miscoordination, highlighting the need for selective communication.
Third, all received messages are treated as equally useful, without mechanisms to filter for information that meaningfully reduces decision uncertainty.
Finally, approaches such as \langcoop are in particular is vulnerable to communication errors or VLM misinterpretations, and lack guarantees for safe decision-making under operational constraints.
To address these challenges, \textbf{we make the following three contributions:}
\begin{enumerate}
    \item \textbf{Efficient and Robust Communication framework:} We propose \textbf{U}ncertainty-Guided \textbf{N}atural Language \textbf{C}ooperative \textbf{A}utonomous \textbf{P}lanning (\oursnospace): a zero-shot, two-stage natural language-based communication and planning framework for CAVs that explicitly incorporates perception uncertainties into decision-making. In the first stage, we introduce \textbf{B}andwidth-\textbf{A}ware \textbf{R}educed \textbf{E}xchange (\barenospace) followed by \textbf{S}elective \textbf{P}rocess for \textbf{A}gent \textbf{R}easoning \textbf{E}xchange (\sparenospace) in the second stage, which together allow an ego CAV to select relevant communication partners in an online fashion, thereby \emph{improving bandwidth efficiency}. Further, we propose a mechanism that enables communicating CAVs to quantify and share their perception uncertainty. The ego CAV then strategically fuses only the most informative messages in a \emph{zero-shot} manner, \emph{improving robustness and safety in cooperative planning}.
    \item \textbf{Accounting for Uncertainty and Mutual Information:} We introduce Information Gain (IG), an uncertainty-guided metric that extends beyond conventional metrics such as driving score and time-to-collision, which do not account for uncertainties in transmitted communication. This metric evaluates the perception and planning uncertainties in VLMs for multi-agent communication and quantifies the value of shared observations via the mutual information between communicating CAVs.
    
    \item \textbf{Enhanced Empirical Performance:} We evaluate \ours on OPV2V \cite{opv2v}, a diverse dataset for CAV driving scenarios in the CARLA simulator \cite{carla}, to demonstrate that \ours reduces bandwidth cost by $63\%$, increases driving score by $31\%$, lowers planning uncertainty by $61\%$, and improves safety with a $4\times$ increase in collision distance margin in near-miss scenarios.
\end{enumerate}
Together, these contributions establish a decision-making framework for VLM-based CAV coordination that is both uncertainty\allowbreak-guided and communication-efficient, and demonstrate measurable improvements in safety and planning performance.

\section{Related Works}
\textbf{Language for Autonomous Driving.} The capabilities of Large Language Models (LLMs) for decision-making are of increasing interest in the field of autonomous driving, especially in the \emph{non-cooperative} setting. Prior non-cooperative works have focused on incorporating (possibly evolving) knowledge bases and common sense human-style reasoning into high-level driving decision-making \cite{mao2023gptdriver,wen2024dilu,mao2024a,sha2025languagempclargelanguagemodels,cui2024large}, and using language as a medium for conflict resolution and negotiation at intersection-like scenarios, where agents have to coordinate the order in which they cross \cite{liu2025colmdriver}.
These approaches remain non-cooperative in nature and primarily address decision-making for a CAV, without leveraging inter-agent communication or coordination.

\textbf{Cooperative Autonomous Driving.} Research in the field of cooperative CAV communication and decision-making has primarily focused on \emph{cooperative perception}, where CAVs share sensor data \cite{gao2018object, chen2019cooper, arnold2020cooperative, cui2022coopernaut, han2023shared} and/or network features \cite{wang2020v2vnet, xu2022v2x,wang2023core}. However, such raw sensor data is often expensive to store, transmit, and infer, and the latency of these operations can be unsuitable for large-scale cooperative driving scenarios. Recent works have investigated the use of text as a medium of communication between CAVs \cite{hu2024agentscodriver, gao2025langcoop, cui2025towards}. However, it is not clear how these works scale beyond the immediate two-CAV scenario, where the questions of ``who to talk to" and ``what information to filter out" naturally arise.

\textbf{Uncertainty Calibration in Decision Making.}
Several post-processing calibration techniques have been proposed to improve the reliability of predictive models. Calibration ensures that the model’s predicted probabilities accurately reflect the true likelihood of correctness.
We focus on methods that quantify uncertainty in classification tasks by estimating the conditional probability that a given sample belongs to each class, conditioned on the observed input.
Platt scaling \cite{platt1999probabilistic} fits a logistic regression to model outputs to produce calibrated probabilities but is primarily restricted for binary classification. For multiclass settings, more expressive methods such as temperature scaling \cite{guo2017calibration} and Dirichlet calibration \cite{kull2019beyond} provide greater flexibility and improved calibration performance. Temperature scaling introduces a single temperature parameter to soften logits, offering a simple calibration method for neural network-based approaches without affecting accuracy, while vector scaling learns class-specific scaling and bias parameters \cite{guo2017calibration}. Dirichlet calibration fits a Dirichlet distribution over class probabilities. Beyond such probabilistic calibration methods, conformal prediction \cite{vovk2005algorithmic, shafer2008tutorial} is a popular approach and offers a model-agnostic framework to construct prediction sets or intervals with probabilistic guarantees; conformal prediction has been extensively utilized in safety-critical autonomous system frameworks \cite{li2024any2anyincompletemultimodalretrieval,bhatt2025knowyoureuncertainplanning, mao2025safeigiveni}.

\section{Methodology}\label{section: methodology}
We present an uncertainty-guided communication and planning approach for VLM-enabled CAVs. Our method is designed to answer three central questions: (i) who to communicate with, (ii) how to quantify uncertainty and the value of communication, and (iii) how to guarantee safety of decisions made under uncertainty.

\begin{figure}[t]
    \centering
    \includegraphics[width=\linewidth, trim={0cm 4.8cm 11cm 0cm}, clip]{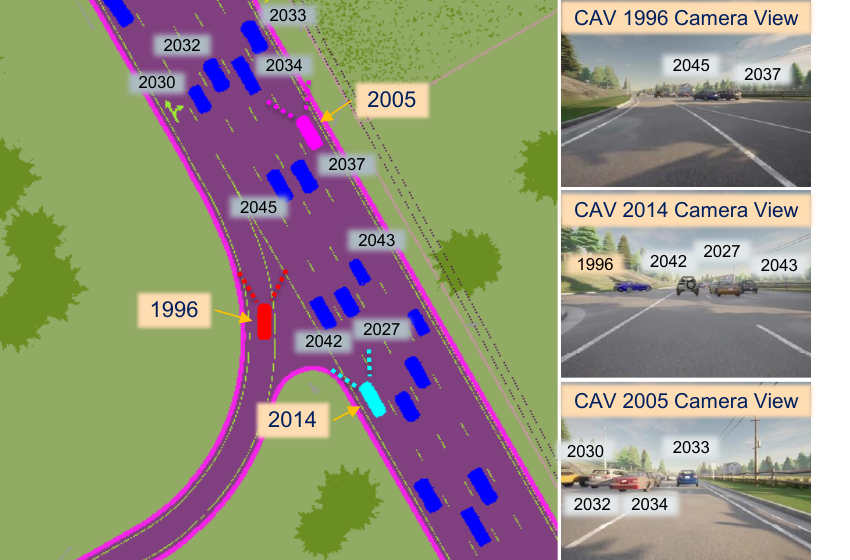}
    \caption{(Left) Labeled bird’s-eye view depicting CAV 1996's intention to merge into highway traffic. CAV 2014 detects relevant vehicles such as 2042 and shares this information with CAV 1996, aiding in decision making. (Right) Front camera views of CAVs.}
    \Description[<short description>]{}
    \label{fig:semantic_bev_with_labels}
    \vspace{-1.75em}
\end{figure}

\textbf{Running Example.} Throughout \Cref{section: methodology}, we use a highway merging scenario as a running example to illustrate our approach. As shown in \Cref{fig:semantic_bev_with_labels}, this scenario involves CAV $1996$ (assumed as ego CAV) merging onto a highway where CAV $2014$ provides relevant information about 2042, which CAV $2005$ cannot.

\subsection{Problem Setup}
We consider a multi-CAV driving scenario involving a set of CAVs $\mathcal{V} = \{v_1, v_2, \dots, v_N\}$ operating in a shared environment. Each vehicle $v_i \in \mathcal{V}$ perceives its surroundings through onboard sensors and obtains a local observation $o_i$. From $o_i$ and onboard sensors, the vehicle generates a textual semantic state representation $s_i = (p_\ego, \dot{p_\ego})$, where $p_\ego$ and $\dot{p_\ego}$ denote its position and velocity, respectively.
The textual representation $s_i$ captures high-level contextual information that may be used for selective communication.
Examples of $s_i$ have been provided in Appendix \ref{app:prompts}.

Given the goal to minimize the communication bandwidth budget while preserving safety, each CAV $v_i$ must decide (i) which peers $v_j \in \mathcal{V} \setminus \{v_i\}$ to communicate with, (ii) what subset of semantic content from $v_j$ is valuable, and (iii) what plan to execute to drive safely.
The objective of all CAVs is to minimize bandwidth consumption while ensuring that they generate a plan above a specified confidence threshold.
We formulate this as a constrained optimization problem:
\begin{equation}
\begin{aligned}
    \min_{\forall b_{ij} \in \{0, 1\}, \pi_i } & \quad \sum_{i,j \in [1,...,N], i \neq j } b_{ij}, \\
    \text{s.t.} & \quad
    \max_{j} b_{ij} I_p(o_j; \pi_i \mid o_i) \geq \tau_\text{safety},  \forall i, 
    \label{eq:problem}
\end{aligned}
\end{equation}
where $b_{ij}$ is a binary integer indicating the bandwidth cost of communicating with $v_j$, $I_p(o_j; \pi_i \mid o_i)$ is the pointwise mutual information between the set of shared observations $o_j$ and the ego CAV’s plan $\pi_i$, representing the reduction in planning uncertainty, and $\tau_\text{safety}$ is a threshold ensuring that the communicated information is sufficient for safe planning.

In realistic deployments, communication bandwidth is not binary as in \Cref{eq:problem}; CAVs can reduce transmission costs through selective and compressed messaging.
For instance, in the proposed method, broadcasting minimal textual descriptions can replace high-bandwidth image sharing while preserving essential context, further reducing communication costs. Also, semantic messages may still suffer from noise or delay.
The formulated problem simplifies this realistic setting, where each vehicle must adaptively prioritize uncertainty-reducing and safety-critical information (e.g., “pedestrian crossing ahead”) while suppressing redundant or low-impact updates (e.g., “lane keeping stable”) to maintain safe and robust planning under dynamic network and sensing conditions.

\subsection{Overview of \oursnospace}
We now describe the overall proposed method, \oursnospace. As shown in \Cref{fig:framework}, \ours has four stages.
In stage \circled{1}, all CAVs $\mathcal{V}=$ \allowbreak $\{v_1, \dots,  v_N\}$ broadcast a lightweight message containing their position and velocity, without sharing their sensor observations.
The message exchange has complexity $O(N^2)$ as it occurs pairwise and is not selective. In the running example, it corresponds to CAVs $1996$, $2005$, and $2014$ sending their positions and velocities to others.

After broadcasting, in stage \circled{2}, each CAV $v_i$ reasons to determine whether a peer CAV $v_j$ provides relevant observations and selects a smaller subset of CAVs to initiate textual communication with. In stage \circled{3}, the ego CAV then evaluates the mutual information between its own observation and each selected CAV $v_j$’s observation, measuring how relevant the shared information is for planning. A higher mutual information value indicates a more useful observation from CAV $v_j$, allowing prioritization of information that best enhances the ego vehicle’s awareness.
The fused messages, observations, and task specifications then serve as input to a VLM in stage \circled{4}, which outputs driving decisions with uncertainty scores to support safe and interpretable planning.

In the running example, CAV $1996$ selects to communicate about semantic information with only CAV $2014$ and not with CAV $2005$, as it has passed beyond the merging area. To arrive at this decision, CAV $1996$ reasons on the mutual information between its observation and CAV $2014$'s observation.

\oursnospace~emphasizes bandwidth efficiency, training-free use, uncertainty\allowbreak-aware decisions, and scalability. Examples of all exchanged messages and VLM prompts are shown in Appendix \ref{app:prompts}.

\subsection{BARE to SPARE: Selective Communication via Natural Language}
For an ego CAV, choosing which CAVs to communicate with is crucial. In general, for $N>2$ vehicles, as $N$ increases, communicating with all CAVs quickly becomes expensive, redundant, and even unsafe (since contributions from irrelevant communicating vehicles can lead to spurious scene understanding). Moreover, pairwise semantic communication, as in \langcoop \cite{gao2025langcoop},~scales poorly with combinatorial dependence. This suggests the clear need for an autonomous vehicle to reason about identifying a \emph{smaller yet relevant subset of vehicles} to communicate with. 
\par
To address this challenge, we develop \textbf{bandwidth-aware reduced exchange} (\barenospace) and \textbf{selective process for CAV reasoning exchange} (\sparenospace).
\barenospace~consists of broadcasting an initial, low-bandwidth package with minimal text-based information from the semantic message that every CAV transmits globally. It consists of the transmitting CAVs position $p_{\cav}$ and heading angle, indicated by their velocity $\dot{p}_\cav$. In the running example, \barenospace~comprises of CAVs $1996,2005$, and $2014$ all communicating their positions and velocities to each other. Following this global broadcast, \sparenospace~enables an ego CAV to efficiently select communication partners  using a heuristic scheme based on intuitive geometric reasoning, thereby improving overall bandwidth efficiency and communication relevance. Specifically, the ego vehicle with position $p_\ego$ and goal position $p_{\goal,\ego}$ for the near future  receives the \barenospace~packet from every CAV, and decides to communicate with a certain CAV if \emph{all} the following conditions are true:
\begin{align}
    \|p_\ego-p_\cav\|_2&\leq d, \label{eq: pego - pcav epsilon}\\
    \left(p_{\goal,\ego} - p_\cav\right)^\top& \dot{p}_\cav > 0,\text{ and} \label{eq: cav correct direction}
\end{align}
where $d>0$ in \Cref{eq: pego - pcav epsilon} represents a ``relevant" distance parameter, that can be set according to the driving environment (\emph{e.g.}, a block distance in urban traffic). \Cref{eq: pego - pcav epsilon,eq: cav correct direction} ensure the ego CAV communicates only with the CAVs which (i) are close enough to it and (ii) are heading towards the ego CAV's goal (and have observations relevant to the ego CAV).

In the running example, these rules will ensure that $1996$ selects to communicate only with $2014$, and $2005$ decides not to communicate with either $1996$ or $2014$, as neither offers it relevant information. These simple geometric arguments allow a CAV to use \bare and \spare to narrow down the set of relevant CAVs to continue communicating with. 
Pseudocode for this stage is presented in \Cref{alg:selective_commu}.
 
\begin{algorithm}[t]
    \caption{Selecting CAVs to communicate with (\barenospace~\& \sparenospace)}
    \label{alg:who to talk to}
    \begin{algorithmic}[1]
        \Require Ego CAV current and goal positions $p_\ego, p_{\goal,\ego}$, set of \barenospace~packets for other CAVs with positions and headings $\mathcal{D}_\barenospace=\{i, p_\cav^i, \dot{p}_\cav^i\}_{i\neq\ego}$, distance threshold $d>0$
        \Ensure Set $\mathcal{C}$ of relevant CAVs to communicate with
        \State Initialize $\mathcal{C}\leftarrow \emptyset$
        \For{$i=1,\dots,|\mathcal{D}_\barenospace|$} \Comment{\sparenospace~process}
        \If{$p_\cav^i, \dot{p}_\cav^i$ satisfies \Cref{eq: pego - pcav epsilon,eq: cav correct direction}}
        \State $\mathcal{C}\leftarrow \mathcal{C}\cup\{i\}$\Comment{Add $i$ to communicating CAV set $\mathcal{C}$}
        \EndIf
        \EndFor\\
        \Return $\mathcal{C}$
    \end{algorithmic}
\label{alg:selective_commu}
\end{algorithm}

Once CAV observations have been selectively communicated, we generate a bird's eye view (BEV) image that captures fused observations with uncertainty quantification and mutual information. An example BEV is presented in \Cref{fig:bev}.

\subsection{Data Fusion via Uncertainty Quantification and Mutual Information}
To determine the value of communication, we introduce metrics grounded in uncertainty quantification and information theory.
In \oursnospace, communication is grounded in perception uncertainty and quantified at the object level. This allows agents to exchange only those object detections that reliably reduce uncertainty in the ego vehicle’s scene understanding.

\textbf{Per-Object Perception Uncertainty.}
Each CAV $v_i$ observes an image $o_i$, yielding a set of detected objects $Y_i = \{y_1, \dots, y_K\}$ and the corresponding raw confidence scores $\{\hat p(y_k \mid o_i)\}_{k=1}^K$ (\textit{i.e.}, class conditional probabilities) obtained from a vision-based detector, such as YOLOv9 \cite{wang2024yolov9} consisting of a projection head applied to VLM embeddings. For each detected object $y_k \in Y_i$, the detector produces a \emph{class-wise confidence vector}
\begin{equation}
\hat{p}(y_k \mid o_i) = [\, \hat{p}(c_1 \mid y_k, o_i), \dots, \hat{p}(c_L \mid y_k, o_i) \,],
\end{equation}
where each entry denotes the probability that the object $y_k$ belongs to class $c_l$ out of $L$ possible classes.
The top-1 predicted class and its corresponding confidence are given by 
$\hat{c}_k = \arg\max_l \hat{p}(c_l \mid y_k, o_i)$ and 
$\hat{p}_{\max}(y_k \mid o_i) = \max_l \hat{p}(c_l \mid y_k, o_i)$, respectively. 
Using \sparenospace, these class-wise confidence vectors are transmitted selectively to the ego CAV. 
These raw confidence scores $\{\hat{p}(y_k \mid o_i)\}_{k=1}^K$ are calibrated using conformal prediction~\cite{vovk2005algorithmic, shafer2008tutorial, angelopoulos2021gentle}, 
providing probabilistic guarantees on these class predictions matching the ground truth labels $Y^*_k$. 
Note, however, that other confidence calibration methods can also be employed, as \emph{\ours is agnostic to the specific calibration technique}. We use conformal prediction in this work solely for the sake of illustration.

This \emph{perception uncertainty score} associated with the raw confidence scores is defined as
\begin{equation}
    u_p(y_k \mid o_i) = 1 - p(y_k \mid o_i),~k=1,\dots,K,
\end{equation}
where $\{p(y_k \mid o_i)\}_{k=1}^K$ denotes the calibrated confidence scores,
The calibration process yields one calibrated uncertainty score per object detection, representing the CAV's uncertainty about its detection of object $y_k$; higher scores indicate greater uncertainty.
We call a confidence score \emph{calibrated} when it reflects the true likelihood of correctness. We now describe the calibration process.

\textbf{Conformal Confidence--Uncertainty Calibration.} The observation $o_i$ is passed through a vision encoder $V$, which operates on the entire image to produce a feature embedding. A projection or detection head $H$ then maps these embeddings to a set of per-object confidence vectors $c_k = H(V(o_i))_{y_k} = [p(c_1|y_k, o_i), \dots, p(c_L|y_k, o_i)$ $]$, where each vector represents the model’s softmax confidence scores over the $L$ object classes for the detected object $y_k$. Given a calibration set $\{(o_i, Y^*_i)\}_{i=1}^N$, where $Y^*_i$ is a set of ground truth labels for the set of objects $Y_i$, we compute nonconformity scores using $S_{nc} = \{ 1 - H(V(o_i))_{y_k} \}_{i=1}^N,$
which quantify how much each prediction deviates from the ground truth. Using $S_{nc}$, we empirically estimate a probability density function $f_{nc}$ from these nonconformity scores.  

Given this distribution and an error tolerance $\epsilon$ on the probability of correctly classifying an object, we can obtain a corresponding calibrated confidence score $c^*$, which according to the theory of conformal prediction, is the $1 - \epsilon$ quantile of $f_{nc}$. Conversely, given $c^*$, we can compute the corresponding error bound as $\epsilon = 1 - \int_{0}^{c^*} f_{nc}(x)\,dx.
$
Given a new image $o_{N+1}$ (not in the calibration set), let $H(V(o_i))_l$ return a confidence vector $c \in \mathbb{R}^L$. Using $f_{nc}$, we can find a \emph{prediction band} for each object $y_k$ $\widehat{C}(o_{N+1}) = \{\, l : H(V(o_{N+1}))_{y_k,l} > 1 - c^* \,\},$
such that the probability of the ground-truth label $y_{N+1}^*$ being contained in $\widehat{C}(o_{N+1})$ is bounded by $1 - \epsilon$.
Formally, the conformal prediction coverage guarantee~\citep{vovk2005algorithmic} ensures $P\big[y_{N+1}^* \in \widehat{C}(o_{N+1})\big] \ge 1 - \epsilon.
$
For downstream planning, we want this prediction band to be a singleton set containing the top-1 class prediction. To enforce this, given a confidence vector $c$, we define $c^* = 1 - \text{sort}(c)_{-2},
$
where $\text{sort}(c)_{-2}$ denotes the second-highest confidence value. Substituting this into the expression for
$\widehat{C}$ yields
\begin{align}
    &\widehat{C}(o_{N+1}) = \{\, l : H(V(o_{N+1}))_l > 1 - c^* \,\} \label{eq:singleton}\\
&= \{\, l : H(V(o_{N+1}))_l > \text{sort}(c)_{-2} \,\} = \{\, \arg\max_l H(V(o_{N+1})_l) \,\}. \notag
\end{align}

This results in a singleton prediction set containing only the most confident class.

By conformal prediction, there exists an $\epsilon \in [0,1]$ such that
\begin{equation*}
P\big[y_{N+1}^* \in \widehat{C}(o_{N+1})\big]
= P\big[y_{N+1}^* = \arg\max H(V(o_{N+1}))\big]
\ge 1 - \epsilon.
\end{equation*}
Thus, the calibrated perception uncertainty, which is the lower bound on the probability of incorrectly identifying an object is expressed as
\[
u_p = \int_{0}^{c^*} f_{nc}(x)\,dx,
\quad \text{where } c^* = 1 - \text{sort}(c)_{-2}.
\]

This formulation provides a rigorous, probabilistically grounded estimate of perception uncertainty, ensuring that
$u_p$ faithfully reflects the likelihood of correctness. %

\textbf{Uncertainty Fusion.}
When ego CAV $v_i$ and neighbor CAV $v_j$ both detect the same object $y_k$, we define the fused probability as
\begin{equation}
    p(y_k \mid o_i, o_j) = \max \big(p(y_k \mid o_i), \, p(y_k \mid o_j)\big).
\end{equation} 
The resulting fused uncertainty is therefore
\begin{equation}
    u_p(y_k \mid o_i, o_j) = 1 - p(y_k \mid o_i, o_j) 
    = \min \big(u_p(y_k \mid o_i), \, u_p(y_k \mid o_j)\big),
\end{equation}
which corresponds to adopting the detections from the least uncertain view.
This uncertainty-fusion method ensures that joint perception is never worse than the best observer, while avoiding overconfidence from a single uncertain calibrated detection.

\textbf{Perception-Based Mutual Information.}  
The utility of incorporating object $y_k$ from CAV $v_j$ is quantified by pointwise mutual information (PMI) \cite{pmi}:  
\begin{equation}
    I_p(y_k ; o_j \mid o_i) = \log \frac{p(y_k \mid o_i, o_j)}{p(y_k \mid o_i)}.
    \label{eq:perception_pmi}
\end{equation}
The fused probability $p(y_k \mid o_i, o_j)$ is computed using the min-uncertainty rule. Objects with high PMI values correspond to cases where a neighbor’s observation reduces ego uncertainty about $y_k$.  
Intuitively, the denominator measures confidence in $y_k$ using only the ego CAV’s observation, while the numerator reflects confidence when both ego and neighbor observations are available. PMI thus represents the \emph{marginal perception confidence gain} provided by CAV $j$.
If $I_{p}(o_j; \pi_i \mid o_i) > 0$, the neighbor contributes positively to ego’s plan confidence; if $I_{p}(o_j; \pi_i \mid o_i) \leq 0$, the contribution is uninformative or harmful.

Communication is performed at the \emph{object level}: for each detection, ego CAV evaluates its calibrated uncertainty score and computes the PMI contribution. Only detections that both reduce uncertainty and yield positive PMI are selected for fusion. The resulting fused object sets are then passed to VLM for planning.

\textbf{Running Example.} In the running example, CAV~$2014$ calibrates the perception confidence of detected objects and transmits the selected results to CAV $1996$ according to \sparenospace. CAV $1996$ then compares its locally observed objects with those reported by CAV $2014$. In this example, the two vehicles observe no common objects, reflecting highest PMI. However, if overlap occurs, CAV $1996$ computes the confidence boost using \Cref{eq:perception_pmi} to improve planning certainty.

\subsection{VLM-Based Planning with Uncertainty}
Unlike perception models that offer real-time inference, given the higher inference times of existing VLMs, we query them for high-level decisions that are consumed by low-level control for real-time planning.
Each VLM output plan is accompanied by a confidence score:
\begin{equation}
    u_d(o, \pi) = -\log p_\vlm(\pi \mid o),
\end{equation}
where $p_\vlm: \mathcal{O} \times \Pi \to [0, 1]$ is the VLM next token likelihood of an output plan $\pi \in \Pi$ given input observation $o \in  \mathcal{O}$. $o$ may represent the ego vehicle’s camera observation and the shared observation from other CAVs, and a natural language prompt to explain the driving scenario, while $\pi$ denotes the corresponding high-level plan, e.g., “wait, then proceed after the vehicle on the right”. This negative log-likelihood serves as an uncertainty measure and may be calibrated using conformal prediction \cite{bhatt2025knowyoureuncertainplanning}.

Similar to perception uncertainty, we adopt a formulation inspired by \cite{chen2025vibe} to compute the PMI between the ego CAV $v_i$’s plan $\pi_i$ and the shared observation from CAV $v_j$:
\begin{equation}
    I_p(\pi_i ; o_j \mid o_i) = \log \frac{p_{\text{VLM}}(\pi_i \mid o_i, o_j)}{p_{\text{VLM}}(\pi_i \mid o_i)}.
    \label{eq:plan_pmi}
\end{equation}
Here, $p_{\text{VLM}}(\pi_i \mid o_i)$ denotes the planning decision based solely on ego’s local observation $o_i$, while $p_{\text{VLM}}(\pi_i \mid o_i, o_j)$ fuses both ego and neighbor observations.

This formulation ensures \emph{decision safety} since $I_p(\pi_i ; o_j \mid o_i)$ represents the \emph{marginal decision confidence gain} provided by CAV $j$.
If $I_p(\pi_i ; o_j \mid o_i) > 0$, the additional observation $o_j$ provides a meaningful confidence gain for the ego CAV’s plan, strengthening the reliability of the decision.
If $I_p(\pi_i ; o_j \mid o_i) \leq 0$, the neighbor CAV's information does not increase or possibly even decreases the ego CAV's confidence, and the neighbor CAV's information is therefore excluded from fusion.
By filtering shared information in this way, we ensure that only uncertainty-reducing and safety-preserving observations influence the ego CAV's final planning actions.

\section{Experimental Results}
We now evaluate \ours in diverse driving scenarios. Our objectives are to: (i) assess the driving quality resulting from using natural language instead of raw images for communication; (ii) examine the effect of selective communication via \bare and \spare on bandwidth savings; (iii) evaluate the reduction in perception and planning uncertainty; (iv) analyze safety performance in terms of distance margin in near-miss events; and (v) test the robustness of \ours to different VLMs.

\begin{figure}[t]
    \centering
    \includegraphics[width=\linewidth, trim={0cm 1.7cm 0cm 0cm}, clip]{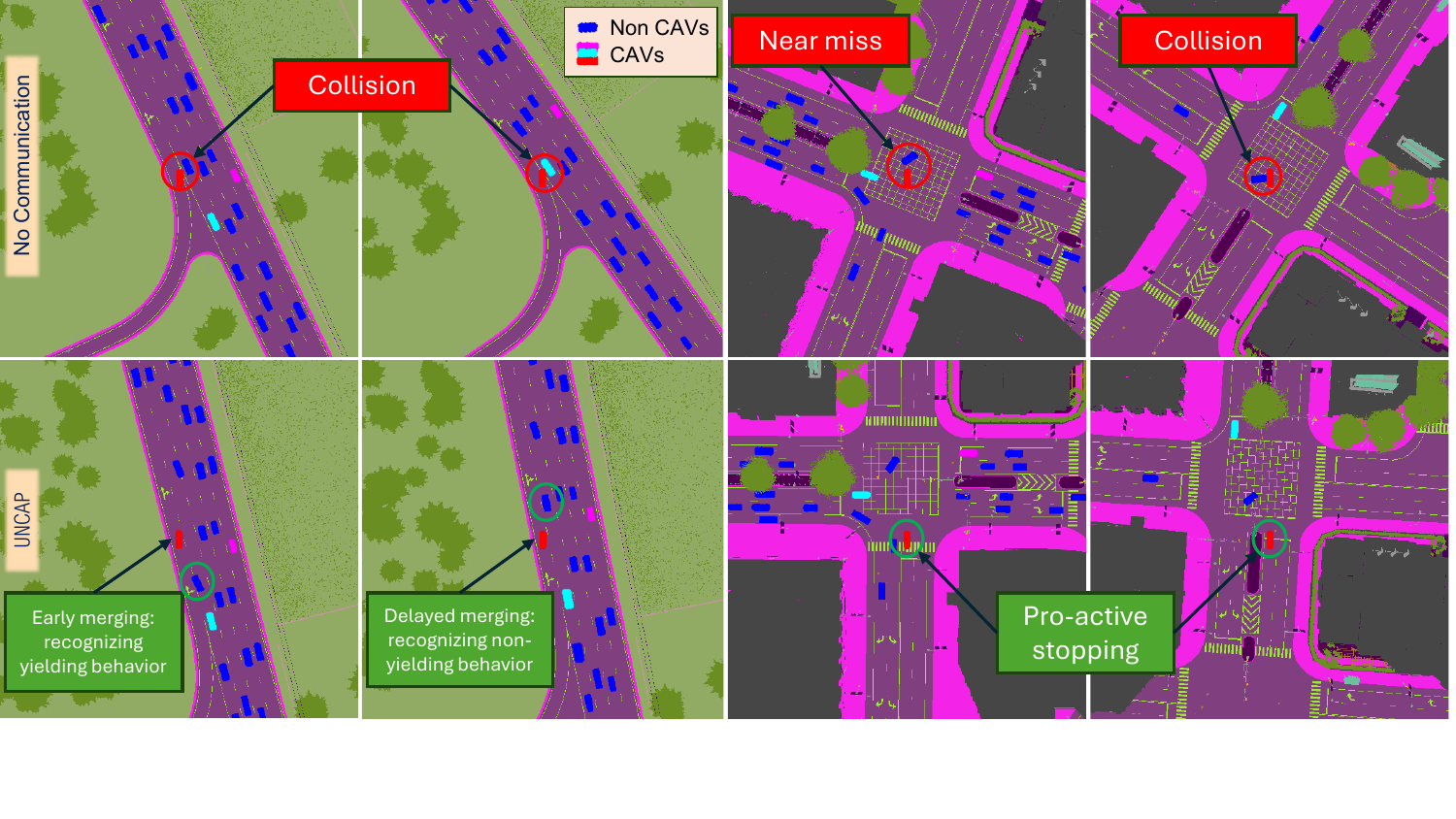}
    \caption{Illustration of 4 scenes featuring highway merging and intersection handling with \nocomm  vs. \oursnospace.}
    \Description[<short description>]{}
    \label{fig:qualitative_results}
\end{figure}

\subsection{Experiment Setup} \label{subsection: exp setup}
We leverage the OPV2V dataset \cite{opv2v}, which is built upon the CARLA simulator \cite{carla}. OPV2V serves as a challenging benchmark specifically curated to evaluate cooperative perception and communication strategies among CAVs. It features diverse, realistic driving scenarios, making it well-suited for testing CAV coordination frameworks such as ours.
We evaluate \ours on several scenarios consisting of 4 diverse challenging layouts from the OPV2V dataset \cite{opv2v}. The scenarios range from highway merging and intersection turning to urban driving with occluded vehicles.
For low-level perception, we use YOLOv9 \cite{wang2024yolov9} in all experiments. For high-level planning, we use GPT-4o \cite{openai2024gpt4ocard}. We also test \ours using different VLMs.
For selective communication in \sparenospace, we set the safety distance threshold $d$ as $50$ meters.
Unless noted otherwise, all methods share identical detection outputs and route definitions to isolate the effect of communication. Perception and communication runs at $10$ Hz. %

\subsection{Evaluation Metrics}
We evaluate \ours using multiple metrics. The \emph{Driving Score} (DS) is a normalized composite of progress, rule compliance, and comfort, with higher values indicating better driving quality. \emph{Route Completion} (RC) measures the percentage of the route completed without failure. The \emph{Infraction Penalty} (IP), starts at $1.0$ and decreases with each traffic violation, down to $0$. DS, RC, and IP are standard CARLA metrics widely used in literature \cite{carlaEvaluationCriteria}. \emph{Total Bandwidth} (TB) records the total communication volume (KB) per episode, where lower values indicate higher efficiency. \emph{Information Gain} (IG) captures the average increase in ego plan confidence (PMI) given received messages. We also analyze inter-vehicle distance margin in near-miss events to evaluate plan safety.

\subsection{Baselines}
We compare \ours against several baselines and present an ablation of all key stages in \ours. \nocomm refers to using only local sensing without any inter-vehicle communication. 
\ours w/o \spare \& Fusion, refers to broadcasting all observations (i.e. without selective communication) without data fusion, and represents existing language-based cooperative planning approaches such as \langcoop~\cite{gao2025langcoop}. 
\ours w/o \spare uses all messages for data fusion but no selective communication. 
\ours w/ Images refers to sharing raw visual inputs in addition to natural language. 
\ours represents the full framework with \barenospace, \sparenospace, and data fusion.

\subsection{Qualitative Results}
\begin{figure}[t] %
    \centering
    \includegraphics[width=0.6\linewidth]{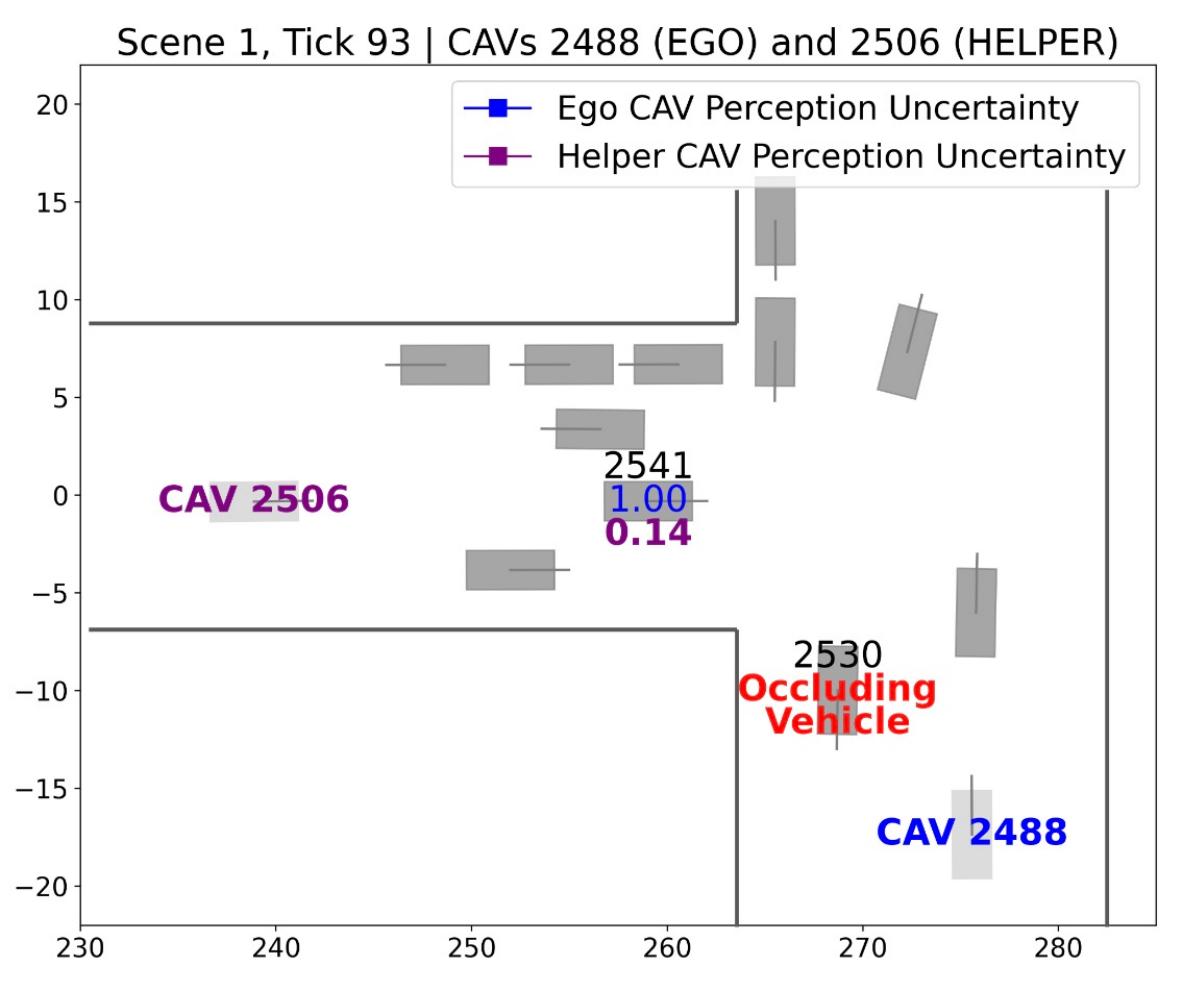}
    \caption{Data fusion illustration. Vehicle 2530 occludes 2541, causing high ego perception uncertainty, which is reduced by fusing the helper CAV's low uncertainty observation.}
    \label{fig:bev}
    \Description[<short description>]{}
\end{figure}

\Cref{fig:qualitative_results} compares highway merging and intersection scenarios. Without communication (\nocomm), ego vehicles rely on local perception, often leading to unsafe maneuvers or collisions due to occlusions. In contrast, \ours enables proactive decision-making: helper CAVs send semantic, uncertainty-calibrated messages via \spare, allowing the ego vehicle to defer merges for fast traffic or enter intersections safely. As shown in the BEV visualizations (\Cref{fig:bev}), \ours uses mutual information-driven fusion to reduce uncertainty for occluded objects (e.g., vehicle $2530$) by leveraging high-confidence detections from selected helpers (e.g., CAV $2506$). By focusing communication on agents directly influencing the ego plan, \ours complements VLM reasoning and significantly reduces overall planning uncertainty.

\begin{figure}
    \centering
    \includegraphics[width=0.8\linewidth]{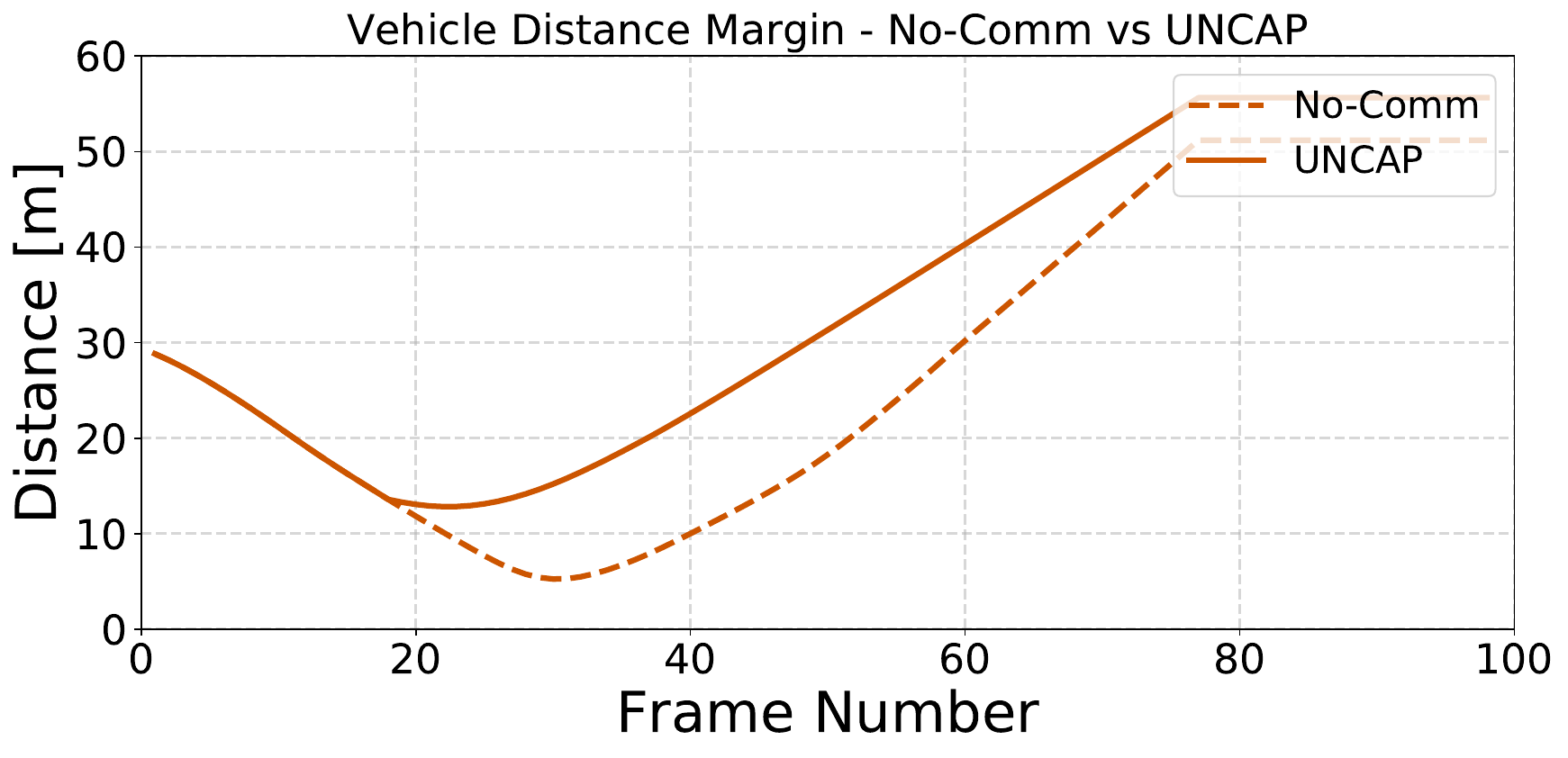}
    \caption{Inter-vehicular distance in the near-miss scenario. Communication enables CAVs to stop within a safe distance.}
    \Description[<short description>]{}
    \label{fig:distance_comm_vs_no_comm}
    \vspace{-1em}
\end{figure}
\begin{figure}
    \centering
    \includegraphics[width=0.8\linewidth]{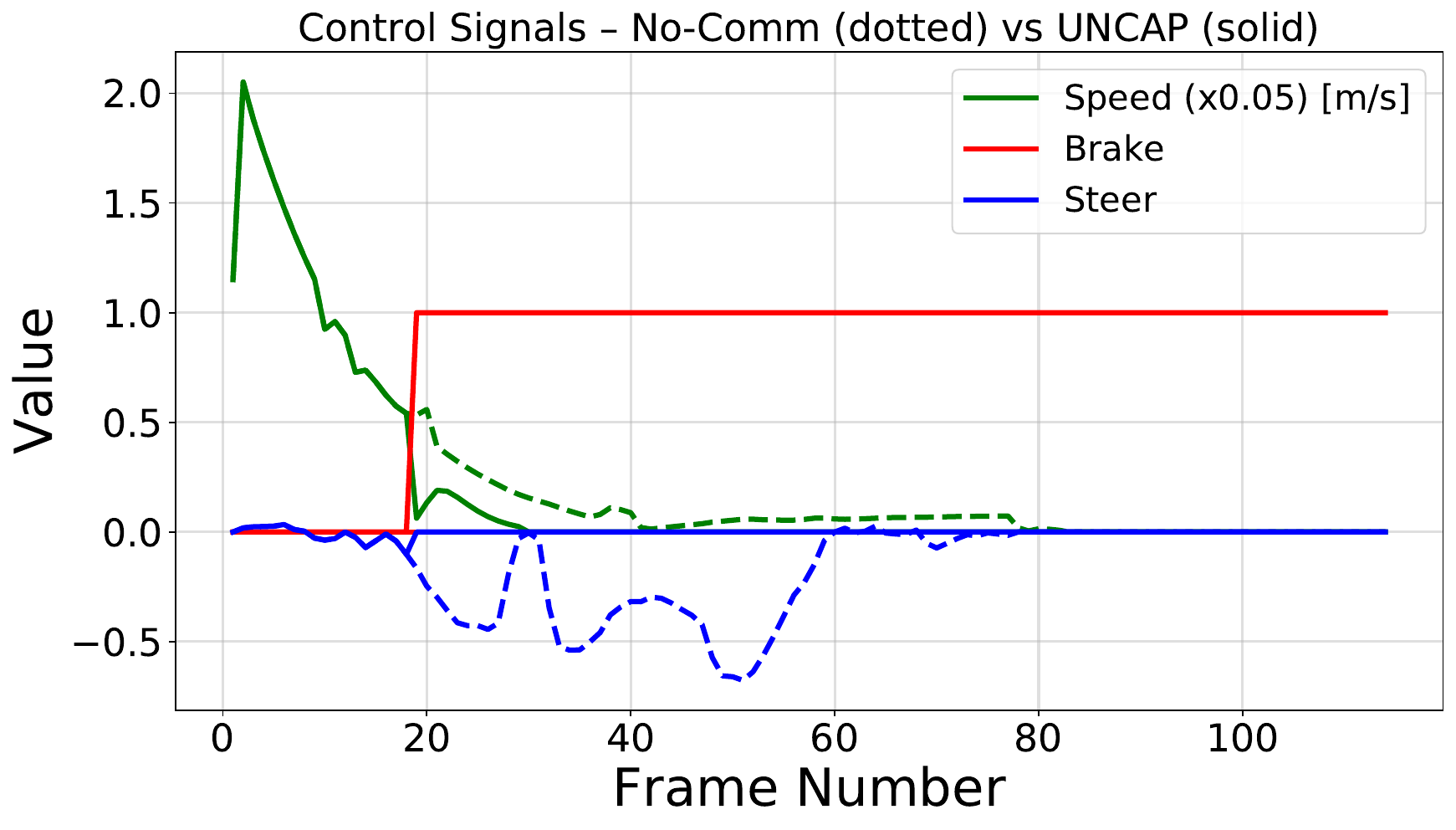}
    \caption{Control signals for the  near-miss scenario.}
    \Description[<short description>]{}
    \label{fig:control_signals_comm_vs_no_comm}
\end{figure}

Across all qualitative cases, \ours consistently exhibits: 
(1) \textbf{Selective awareness:} only uncertainty-reducing detections are fused; 
(2) \textbf{Proactive planning:} ego CAVs anticipate conflicts several frames earlier and initiates proactive breaking, maintaining a higher inter-vehicle distance margin. We illustrate these attributes via \Cref{fig:distance_comm_vs_no_comm} and \ref{fig:control_signals_comm_vs_no_comm}; and 
(3) \textbf{Interpretable cooperation:} messages remain human-readable and auditable.
Overall, these results illustrate that uncertainty\allowbreak-guided natural language communication yields interpretable, bandwidth-efficient, and safer cooperative behavior compared to unstructured or purely visual exchanges.

\subsection{Quantitative Results}
\noindent\textbf{\ours Enhances Driving Quality Compared to Baselines.}
Across diverse scenarios (merges, intersections, occlusions), \ours (\barenospace + \sparenospace) consistently yields superior decision quality. As shown in \Cref{tab:vehicle1_results}, \ours achieves the highest driving score, exceeding \nocomm~by $31.4\%$ and naive broadcasting (i.e., \langcoop~\cite{gao2025langcoop}) by $27.9\%$. Route completion rates show a similar trend, with \ours outperforming all baselines, and it attains the best IP score, indicating the fewest traffic violations.
\begin{table}[t]
    \centering
    \caption{Performance metrics for different communication strategies. \ours outperforms all baselines across all metrics.}
    \begin{tabular}{lccc|c}
    \hline
    Method & DS $\uparrow$ & RC $\uparrow$ & IP$\uparrow$ & TB (KB)$\downarrow$ \\
    \hline
    \nocomm & 48.9\% & 39.7\% & 62\% & -- \\
    \makecell[l]{\ours w/o \spare \& Fusion \\ (represents \langcoop \cite{gao2025langcoop})} & 52.4\% & 79.6\% & 65\% & 89 \\
    \ours w/o \spare & 78.8\% & 87.2\% & 90\% & 89 \\
    \ours w/ Images              & 69.5\% & 88.3\% & 78\% & 33,600 \\
    \ours & \textbf{80.3\%} & \textbf{89.2\%} & \textbf{90\%} & \textbf{33} \\
    \hline
    \end{tabular}
    \label{tab:vehicle1_results}
\end{table}

Notably, the use of image observations in communication degrades planning performance, likely because VLMs struggle to integrate multiple visual inputs, often confusing scene content across views. Bearing this in mind, \ours avoids this pitfall by leverages language-only communication, translating one image per CAV into text locally, which can be seen as a reasoning step. We present further results for \ours on a total of $9$ scenarios from both OPV2V and a custom dataset, which have $2$ to $4$ CAVs in each scenario in Appendix \ref{app: 9 driving scenes}.

\begin{figure}[t]
    \centering
    \includegraphics[width=0.75\linewidth]{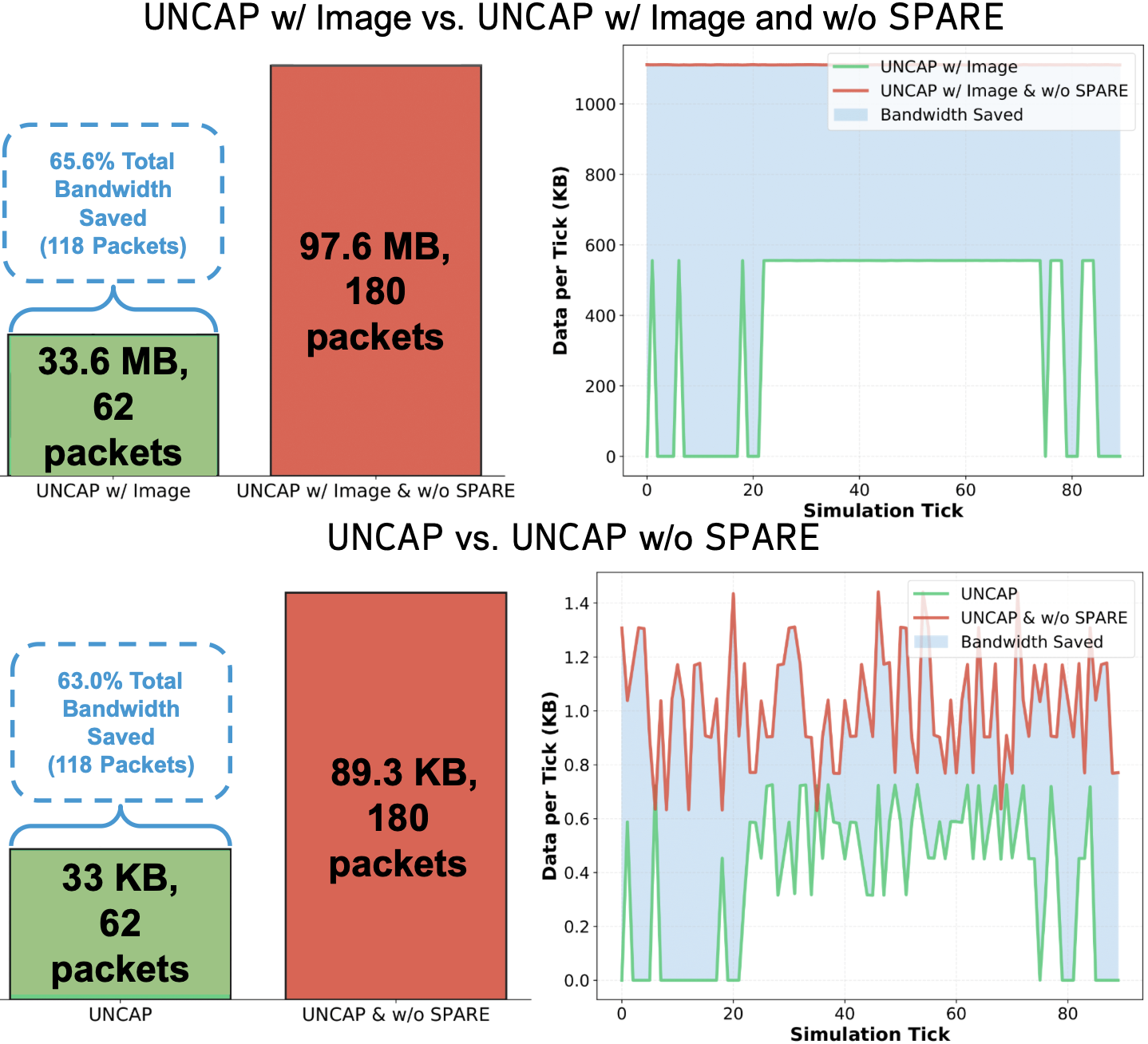}
    \caption{\spare enables substantial bandwidth savings in \ours, whether the communication uses images (top) or text (bottom). 
    (Top) Since we selectively communicate with other CAVs, the bandwidth fluctuates per tick.
    (Bottom) The bandwidth fluctuates as the text messages vary in length.}
    \Description[<short description>]{}
    \label{fig:combined_bandwidth}
\end{figure}

\textbf{\spare in \ours Boosts Bandwidth Efficiency Compared to Baselines.}
A key motivation for language-based communication methods such as \langcoop~\cite{gao2025langcoop} is their substantial bandwidth savings compared to approaches that exchange images or raw sensor data. As shown in \Cref{tab:vehicle1_results} and \Cref{fig:combined_bandwidth}, the \emph{\barenospace~and \sparenospace~processes enable \ours to achieve state-of-the-art bandwidth efficiency}, requiring only $37\%$ of the bandwidth used by \langcoop~(equivalent to \ours without \sparenospace~and fusion) when using textual communication alone, and only $34\%$ when using both text and image communication.
It underscores the importance of selective communication done in \ours and reveals the poor scalability of conventional frameworks that broadcast all information indiscriminately.

\begin{table}[t]
    \centering
    \caption{Model confidence and information gain across communication methods. IG scores are presented on log scale.}
    \begin{tabular}{lcccc}
        \hline
        \multirow{2}{*}{Method} & \multicolumn{2}{c}{Perception} & \multicolumn{2}{c}{Decision} \\
        \cline{2-5}
         & Confidence & IG & Confidence & IG \\
        \hline
        \nocomm & 0.25 & --- & 0.43 & --- \\
        \ours w/o \sparenospace & 0.71 & 1.04 & 0.44 & 0.02 \\
        \ours w/ Images & 0.71 & 1.04 & 0.48 & 0.11 \\
        \oursnospace & 0.71 & 1.04 & \textbf{0.78} & \textbf{0.60} \\
        \hline
    \end{tabular}
    \label{tab:pmi}
\end{table}

\textbf{\ours Enhances Safety via Uncertainty Quantification and Mutual Information.}
Per Equations (\ref{eq:perception_pmi}) and (\ref{eq:plan_pmi}), PMI measures IG. A positive IG indicates boosted confidence from shared data, while negative values suggest noise. \ours achieves a significant confidence boost of $0.60$. In perception, \ours matches the IG of full broadcasting because our selective fusion identifies the most informative objects without redundant data. For planning, broadcasting without selection introduces noise from irrelevant CAVs, resulting in marginal IG gains over no communication. Furthermore, image-based exchange drops planning IG to $0.11$, confirming that multi-view images are less scalable for VLMs. In contrast, \oursnospace's selective language-based communication reaches an IG of $0.60$, demonstrating superior decision-making confidence and safety.

\textbf{\ours Enables Real-Time Planning.}
A major concern for VLM-based planning is latency, and we verify that \ours makes real-time planning feasible. 
In \oursnospace, VLM is queried only at critical points, when intentions change, and actions must be executed; it avoids the high latency that querying the VLM at every tick would incur.
Communication with KB-scale messages achieves $\sim200$ms transmission time, with an estimated broadcast speed of $1.05$Mbps, a groupcast speed of $1.52$Mbps \cite{an2023enhancing}, and $10$ms overhead \cite{coll2022end}. In contrast, communicating with images incurs higher latency, increasing total transmission time by $1000\times$ to around $200$s, even with selective communication. \ours achieves an average VLM planning latency of $1.33$s, with an overall latency of $\sim1.5$s per decision step. 

\begin{table}[t]
    \centering
    \caption{Comparison of different VLMs across driving and safety metrics, showing \oursnospace’s robustness to model choice. We omit IG for GPT-5 as the API does not provide access to token probabilities.}
    \begin{tabular}{lccc|c}
    \hline
    Model   & DS$\uparrow$ & RC$\uparrow$ & IP$\uparrow$ & IG$\uparrow$\\
    \hline
    GPT-4o-mini    & 64.7\% & 88.2\% & 73\% & 0.42 \\
    GPT-4o   & \textbf{80.3\%} & \textbf{89.2\%} & \textbf{90\%} & \textbf{0.60}\\
    GPT-4.1  & \textbf{80.3\%} & \textbf{89.2\%} & \textbf{90\%} & 0.40\\
    GPT-5    & 74.5\% & \textbf{89.2\%} & 83\% & --- \\
    \hline
    \end{tabular}
    \label{tab:model_comparison}
\end{table}

\textbf{\ours is Robust to The Underlying VLM.}
\Cref{tab:model_comparison} evaluates \ours across various VLMs. The framework maintains consistent performance regardless of the specific model, with larger VLMs like GPT-4o achieving peak scores and smaller models like GPT-4o-mini remaining highly competitive. While Information Gain (IG) varies based on model-specific confidence scores, the consistently positive trend confirms that our communication and fusion framework—rather than a specific VLM—is the primary driver of performance. This highlights the adaptability and robustness of \ours to different VLM architectures.

\section{Conclusion}

We propose \oursnospace, a vision–language model-based planning approach for CAVs using natural language communication. Unlike traditional methods that broadcast raw data or unfiltered text, \ours employs a two-stage process: (i) identifying relevant CAVs and (ii) exchanging messages containing explicit quantitative perception uncertainty. By fusing only the most informative messages, \ours achieves state-of-the-art driving performance and bandwidth efficiency. Our results demonstrate \oursnospace's potential for reliable, scalable planning in complex environments.

\textbf{Future Work.} We plan to extend \ours to mixed-traffic scenarios involving human-driven vehicles and natural language instructions. Additionally, we will investigate the impact of realistic network conditions (e.g., packet loss) and evaluate performance on physical CAV fleets to account for real-world sensor noise and vehicle dynamics.

\begin{acks}
This material is based upon work supported in part by 
the Lockheed Martin Corporation under grant No. MRA16-005-RPP023,
the Office of Naval Research (ONR) under Grant No. N00014-22-1-2254,
the Defense Advanced Research Projects Agency (DARPA) contract FA8750-23-C-1018, DARPA ANSR: RTXCW2231110,
and the Army Research Laboratory under Cooperative Agreement Number W911NF-25-2-0021, W911NF-24-1-0220, and W911NF-23-2-0011.

\end{acks}

\clearpage
\bibliographystyle{ACM-Reference-Format} 
\bibliography{refs}

\clearpage
\appendix
\onecolumn

\section{Examples of Package, VLM Input Prompts, and Output Formats}
\label{app:prompts}

Using the running example in Figure \ref{fig:semantic_bev_with_labels}, we now provide examples of all the broadcasted messages. We also provide the input prompt and output examples we used to query VLMs in our experiments.

\subsection{Broadcast Semantic Message of \barenospace}
The broadcasted semantic message includes each CAV’s absolute position and heading angle (yaw in radians).
\begin{tcolorbox}
    [colframe=datasetAcolor!70!black, colback=datasetAcolor!10!white, colbacktitle=datasetAcolor!70!black,
    coltitle=white,
    title=\textbf{Broadcast Message of \barenospace}, 
    sharp corners=south, boxrule=0.8mm, width=\textwidth, enlarge left by=0mm, enlarge
    right by=0mm] 
    \footnotesize 
[An example message from CAV 2014] `2014': \{'position': [-222.67, 240.10], 'heading': 0.54\}
\end{tcolorbox}

\subsection{Post-\spare BEV Generation}
After receiving the broadcast of \barenospace, each CAV then queries the selected CAVs from \spare to share all their observed objects. For instance, the helper CAV $2014$ provides its observed vehicles $2042$, which CAV $1996$ cannot see directly, to CAV $1996$:
\begin{tcolorbox}
    [colframe=datasetAcolor!70!black, colback=datasetAcolor!10!white, colbacktitle=datasetAcolor!70!black,
    coltitle=white,
    title=\textbf{Shared Observations from Helper CAV}, 
    sharp corners=south, boxrule=0.8mm, width=\textwidth, enlarge left by=0mm, enlarge
    right by=0mm] 
    \footnotesize 
`2042': \{
    angle: 7.70437327446416e-05, -0.108062744140625, 0.16955943405628204
    
    center: -0.057383179664611816, 0.00015066092601045966, 0.9351239800453186

    extent: 2.3022549152374268, 0.9657965898513794, 0.9274230599403381
    
    location: -209.55389404296875, 240.1245574951172, 0.010594921186566353
    
    speed: 34.487769259146035
    
    confidence: 0.8112537264823914
\} ... [other observed objects from CAV 2014]
\end{tcolorbox}

\subsection{VLM High-level Planning Input Prompts and Output Formats}
The ego CAV then inputs its generated BEV and its front-camera view to its local VLM to fuse the observations.
\begin{tcolorbox}
    [ colframe=datasetBcolor!70!black, colback=datasetBcolor!10!white, colbacktitle=datasetBcolor!70!black,
    coltitle=white, 
    title=\textbf{VLM Perception Prompts}, 
    sharp corners=south, boxrule=0.8mm, width=\textwidth, enlarge left by=0mm, enlarge
    right by=0mm ] 
    \footnotesize 
You are an AI assistant that helps with safe driving from a high-level perspective.
You are working with a scenario in which there are some autonomous cars and many regular cars.
You must refer to them by their IDs, which are used to label them.
You are provided with two images:
\begin{itemize}
    \item a birds-eye view of an intersection with some autonomous cars and some regular cars; (see Figure \ref{fig:bev} for example)
    \item the front view of Vehicle \textcolor{blue}{ego\_cav\_id}, called the Ego CAV. (see Figure \ref{fig:semantic_bev_with_labels} for example)
\end{itemize}
In the birds-eye view, the autonomous cars are colored pink and the regular cars are colored yellow.
You must refer to them by their IDs, which are used to label them.
Directions on this map are given as you see them: North is up, South is down, East is right, West is left.
The vehicle of interest in this scenario is Vehicle \textcolor{blue}{ego\_cav\_id}, called the Ego CAV. It currently \textcolor{blue}{ego\_intention}. It is currently facing north.
Your task is to discern which vehicles might interfere with the motion of the Ego CAV such that it should know about them in order to make a safe decision.
At the end of your response, you must include a space-separated list of the vehicle IDs of interest in this EXACT format:
id\_1 id\_2, ... id\_n.
Or, only if there are no vehicle IDs of interest, include at the end of your response the number: 0.
\end{tcolorbox}
The VLM then outputs:
\begin{tcolorbox}
    [colframe=datasetCcolor!70!black, colback=datasetCcolor!10!white, colbacktitle=datasetCcolor!70!black,
    coltitle=white, 
    title=\textbf{VLM Perception Output Example}, 
    sharp corners=south, boxrule=0.8mm, width=\textwidth, enlarge left by=0mm, enlarge
    right by=0mm] 
    \footnotesize 
    [Reasoning outputs are skipped here] Relevant vehicle IDs: 2042, 2014, 2027.
\end{tcolorbox}

After the VLMs on each CAV select the relevant IDs using the prompts above, the following semantic message is used for communication.
All the vehicles below are only seen by CAV 2014, and the information is transmitted to CAV 1996, allowing it to plan safely.
\begin{tcolorbox}
    [colframe=datasetBcolor!70!black, colback=datasetBcolor!10!white, colbacktitle=datasetBcolor!70!black,
    coltitle=white, 
    title=\textbf{VLM Structured Input Semantic Message Example for Planning},
    sharp corners=south, boxrule=0.8mm, width=\textwidth, enlarge left by=0mm, enlarge
    right by=0mm] 
    \footnotesize
Ego Vehicle: Facing E, Speed: 36.250440788922994

Vehicle 2042 (perception confidence 0.76/uncertainty 0.24): Relative direction to Ego CAV: SSE, Distance: 6.371047022893454 (close), Facing N, Speed: fast - NOTE: This vehicle is in an adjacent lane

Vehicle 2014 (perception confidence 1.00/uncertainty 0.0): Relative direction to Ego CAV: S, Distance: 18.8218177232921 (far), Facing N, Speed: fast - NOTE: This vehicle is in an adjacent lane

Vehicle 2027 (perception confidence 0.90/uncertainty 0.1): Relative direction to Ego CAV: ESE, Distance: 8.255774681667308 (close), Facing N, Speed: fast
\end{tcolorbox}

\begin{tcolorbox}
    [colframe=datasetBcolor!70!black, colback=datasetBcolor!10!white, colbacktitle=datasetBcolor!70!black,
    coltitle=white,
    title=\textbf{VLM Planning Prompts (Highway Merging Running Example)}, 
    sharp corners=south, boxrule=0.8mm, width=\textwidth, enlarge left by=0mm, enlarge
    right by=0mm] 
    \footnotesize 
Here is the situational description from the perspective of the Ego CAV: \textcolor{blue}{ego\_description}
If 0 descriptions of other cars are provided, don't merge. MERGE DECISION RULES:
\begin{enumerate}
    \item Merge only if the right lane is open.  
    \item Do not merge if a vehicle approaches in the right lane.  
    \item Merge safely if a vehicle in the right lane is behind the ego vehicle and its distance $>$ 10.  
    \item Do not merge if any vehicle in the right lane is closer than 10 units.  
    \item Account for vehicle speed: faster vehicles require more clearance.  
\end{enumerate}
Do not be overly safe. If you see clearance over 10 distance you have clearance to merge.
Analyze the relative positions, distances, and speeds of vehicles.
Respond strictly in this format:
action: \textcolor{red}{[merge|no merge]}
reason: \textcolor{red}{[brief explanation of decision based on vehicle positions and distances]}
\end{tcolorbox}

Lastly, the VLM outputs a high-level driving plan, such as the example below.
\begin{tcolorbox}
    [colframe=datasetCcolor!70!black, colback=datasetCcolor!10!white, colbacktitle=datasetCcolor!70!black,
    coltitle=white, 
    title=\textbf{VLM Plan Output Example}, 
    sharp corners=south, boxrule=0.8mm, width=\textwidth, enlarge left by=0mm, enlarge
    right by=0mm] 
    \footnotesize 
    Action: no merge
    
    Reason: No descriptions are provided about other cars, so the decision is to not merge.
    
    Probability: 0.992902
\end{tcolorbox}

\section{Additional BEV Examples}

\begin{figure}[h]
    \centering
    \begin{minipage}{0.33\linewidth}
        \centering
        \includegraphics[width=\linewidth]{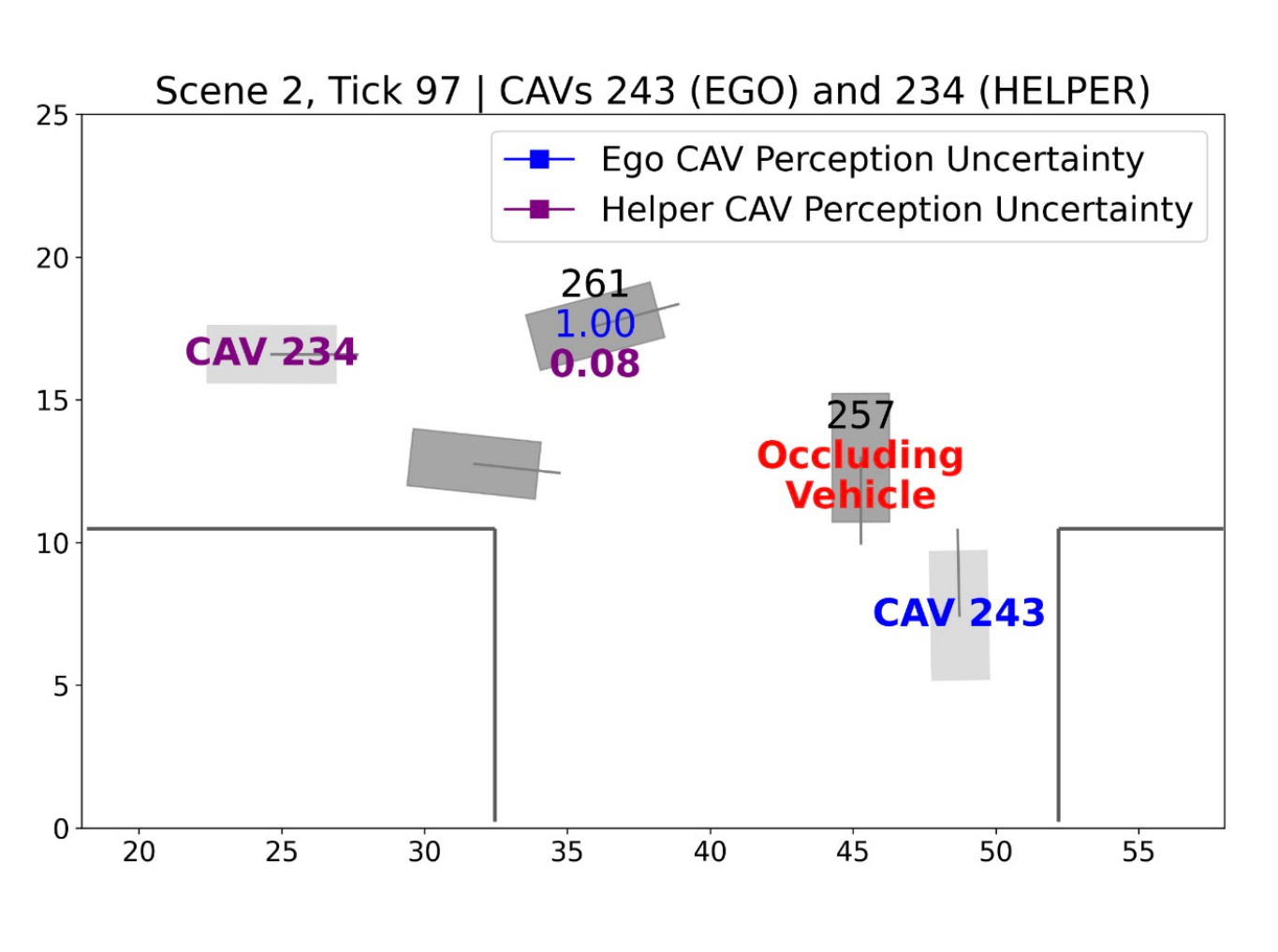}
    \end{minipage}%
    \begin{minipage}{0.33\linewidth}
        \centering
        \includegraphics[width=\linewidth]{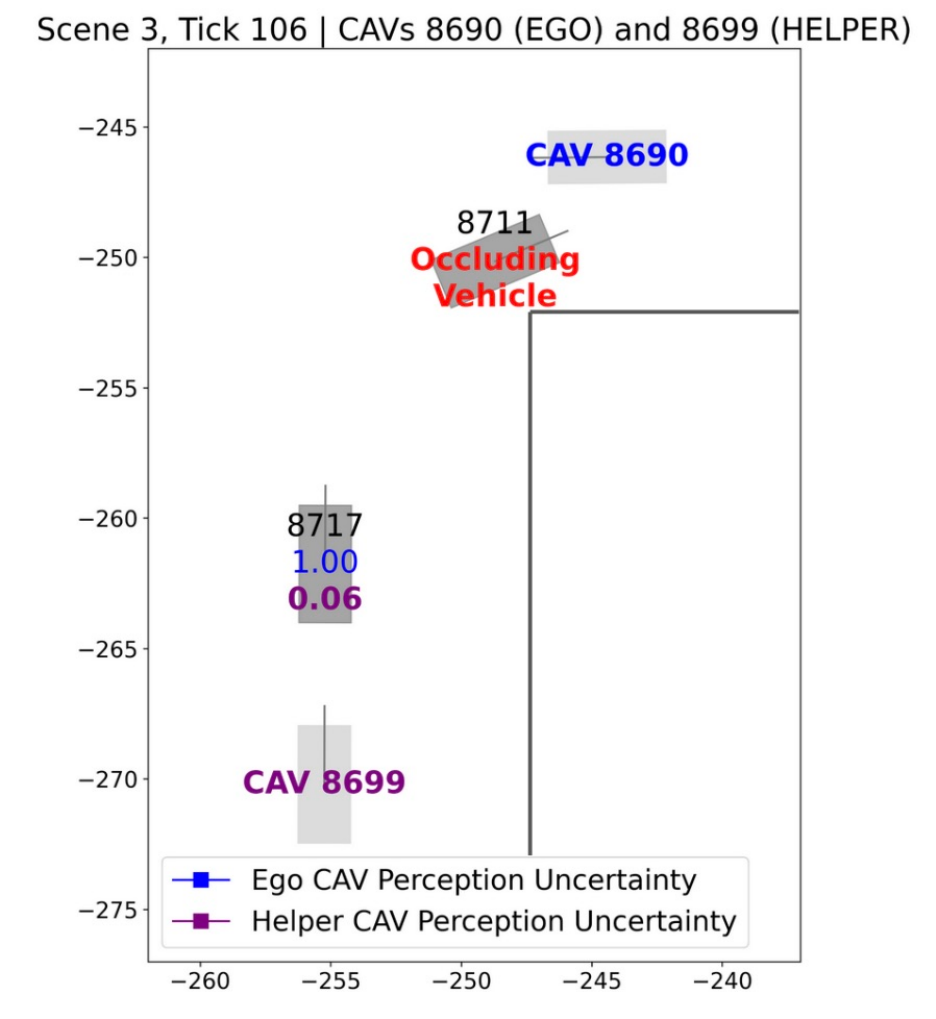}
    \end{minipage}%
    \begin{minipage}{0.33\linewidth}
        \centering
        \includegraphics[width=\linewidth]{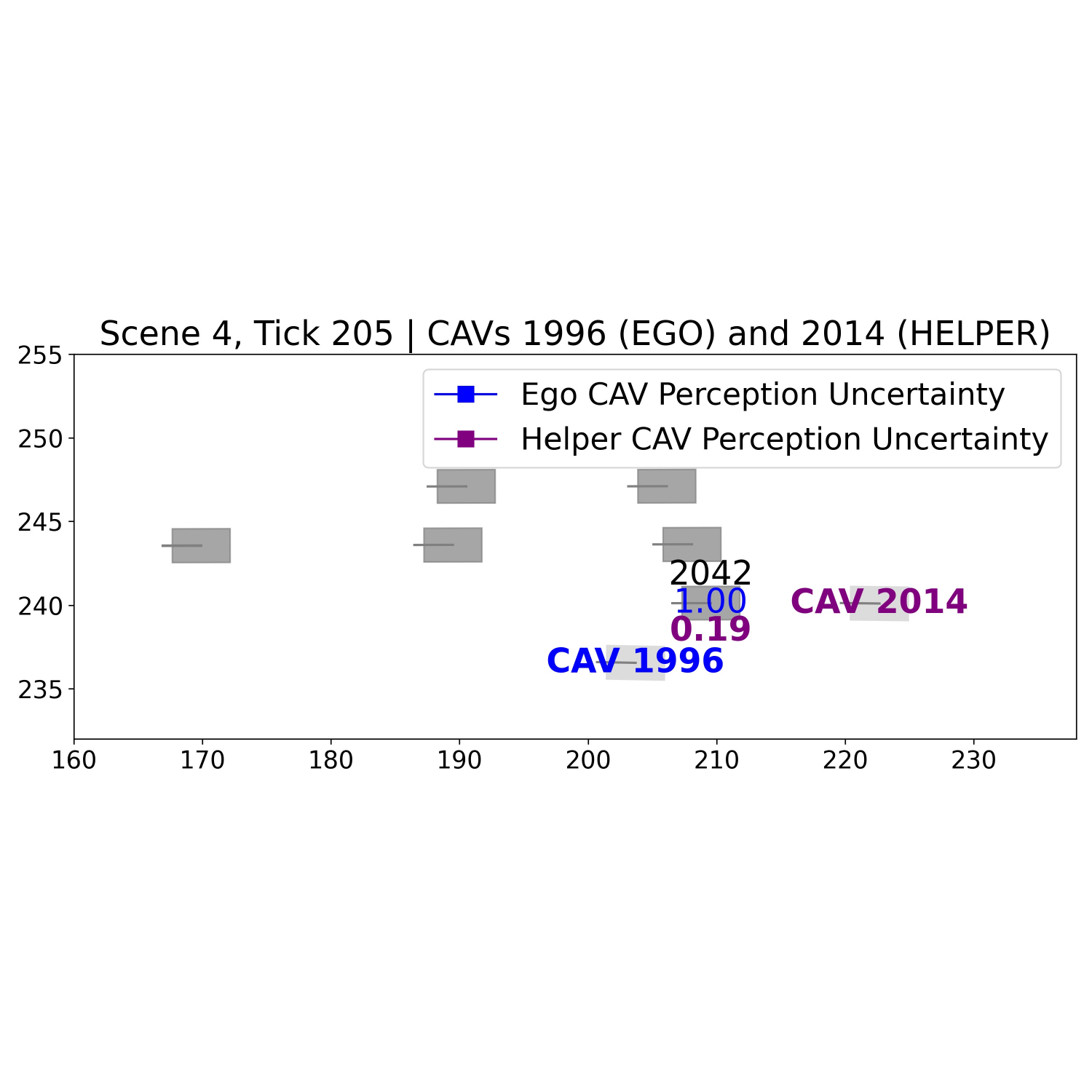}
    \end{minipage}
    \caption{BEVs generated for three selected scenes. In each scene, a vehicle that could potentially interfere with the ego vehicle is identified and communicated to the ego vehicle by a helper CAV that observes the vehicle with lower uncertainty.}
    \Description[Additional BEV examples showing cooperative perception in multi-vehicle scenes.]{Three BEV visualizations showing how helper CAVs communicate observations of interfering vehicles to the ego vehicle, reducing uncertainty and improving decision safety.}
    \label{fig:bev_scene_3}
\end{figure}

Figure~\ref{fig:bev_scene_3} presents three additional representative BEV visualizations that illustrate how cooperative perception reduces uncertainty in complex traffic scenes. Each example shows the ego vehicle receiving critical observations from nearby CAVs that detect potentially interfering vehicles. These shared detections enhance situational awareness and support safer decision-making.

\section{\ours results for more driving scenarios} \label{app: 9 driving scenes}
\Cref{tab:vehicle1_results} shows results for \ours averaged over the 4 challenging scenarios presented in \Cref{fig:qualitative_results}. For the complete set of 9 scenarios mentioned in \Cref{subsection: exp setup}, \ours shows an average driving score (DS) of $84.16\%$, an average route completion (RC) of $89.11\%$, and an average infraction penalty (IP) of $94\%$. We reiterate that a higher value is better for all three metrics.

\end{document}